\documentclass[10pt]{article}

\newif\ifreview
\reviewfalse
\ifreview
  \usepackage{tmlr}
\else
  \usepackage[preprint]{tmlr}
\fi
\ifreview
  \newcommand{\leanwhere}{supplementary material}
\else
  \newcommand{\leanwhere}{ancillary files of this arXiv submission}
\fi

\usepackage{amsmath,amssymb,amsfonts}
\usepackage{mathtools}
\usepackage{amsthm}
\usepackage{graphicx}
\usepackage{booktabs}
\usepackage{xcolor}
\usepackage{url}
\usepackage{hyperref}

\newcommand{\bm}[1]{{\mbox{\boldmath $#1$}}}
\newcommand{\argmax}{\mathop{\rm arg~max}\limits}

\DeclareMathOperator{\artanh}{artanh}
\newcommand{\vpi}{\bm{\pi}}

\newtheoremstyle{nonitalic}%
  {3pt}{3pt}{\normalfont}{}{\bfseries}{.}{.5em}{}
\theoremstyle{nonitalic}
\newtheorem{theorem}{Theorem}
\newtheorem{lemma}{Lemma}
\newtheorem{corollary}{Corollary}

\newtheorem{definition}{Definition}
\newtheorem{remark}{Remark}

\newtheorem{proposition}{Proposition}

\title{When Is Product Decoding of ECOC MAP-Consistent?\\Pair Structure and the $n$-vs-Rest Codes}

\author{%
  \name Gendo Kumoi \email kumoi@vos.nagaokaut.ac.jp \\
  \addr Department of Information and Management Systems Engineering\\
  Nagaoka University of Technology
}

\def\month{MM}
\def\year{YYYY}
\def\openreview{\url{https://openreview.net/forum?id=XXXX}}

\begin{document}

\maketitle

\begin{abstract}
Error-Correcting Output Codes (ECOC) combine binary classifiers by the \textbf{sum decoder}, which totals probabilities, or the \textbf{product decoder}, which totals log-likelihoods. In the \textbf{noiseless setting} $\bm p=C^\top\vpi$, where the binary classifiers output the true posteriors, we ask on which code matrices $C$ each decoder always agrees with the Bayes-optimal (MAP) class. For the sum decoder the answer is known: the code must be equidistant \citep{james1998error}. We answer the question for the product decoder.

The key observation is that two classes are compared through the \textbf{votes} of the columns on which they differ: the sum decoder decides by their \textbf{sample mean}, the product decoder by an \textbf{M-estimator} with influence function $\artanh$, and for every code each vote is the MAP margin plus a bias determined by the background classes alone. Our main theorem states that the product decoder is MAP-consistent if and only if, for every class pair, the columns on which the two classes differ are matched with the columns obtained by swapping the two classes (or their complements)---\textbf{pair structure}---and, equivalently, if and only if the code is a \textbf{union of $n$-vs-Rest codes}. The sum decoder needs only the first moment of the votes to balance; the product decoder needs the whole vote distribution to be symmetric, and this forces the code to be invariant under every permutation of the classes. Efficient equidistant codes are therefore excluded: Hadamard codes ($M\geq8$), which meet the Plotkin bound with $M-1$ columns, have no pair structure at all, so for every type of Hadamard matrix the product decoder can disagree with MAP on every class pair, and for the Sylvester type the reversal margin can be at least a constant $\bar\delta\approx0.0182$ independent of $M$. All theorems are machine-checked in Lean~4.

Numerical experiments confirm these results. With estimated probabilities on six real datasets, the comparison depends on calibration: when each column is calibrated, the product decoder is slightly more accurate (as a sum of log-odds, it is invariant to a temperature common to all columns); when the calibration error differs across columns, the sum decoder, which uses bounded votes, is better. In either case the two decoders disagree only near ties.
\end{abstract}

\section{Introduction}\label{sec:intro}

Error-Correcting Output Codes (ECOC) reduce multi-class classification to binary problems \citep{dietterich1995solving,allwein2000reducing}. Each class $m$ is assigned a code word $\bm{c}_m\in\{0,1\}^n$, and a binary classifier is trained for each column $j$; decoding aggregates the outputs $p_j$ and predicts the class with the ``closest'' code word.

Decoders fall into two groups: the \textbf{sum decoder} ($L_1$ decoding \citep{dietterich1995solving}), which adds the column probabilities, and the \textbf{product decoder} (loss-based decoding \citep{allwein2000reducing,passerini2004new}), which adds the Bernoulli log-likelihoods of the columns. Which should be used? Our answer:

\begin{quote}
\emph{If the classifiers output the true posteriors, the sum decoder always returns the MAP class if and only if the code is equidistant \citep{james1998error}, and the product decoder if and only if the code is a union of $n$-vs-Rest codes. Short codes with large distance, such as Hadamard codes, are equidistant but not of this form, so on them only the sum decoder is safe. For real classifiers with estimation error, the product decoder is slightly more accurate when each column is calibrated individually, and the sum decoder is better when the calibration error varies across columns. In either case the two decoders differ only near ties.}
\end{quote}

We derive the first part of this answer in the \textbf{noiseless setting}, in which the binary classifiers are perfect:
\begin{equation}\label{eq:oracle}
  \bm{p} = C^\top\vpi \in [0,1]^n,
\end{equation}
where $\vpi\in\Delta^{M-1}$ is the true posterior vector. Here any disagreement with MAP is caused by the decoding rule and the code structure, not by estimation error.

The answer matters because codes differ greatly in cost. The number of binary classifiers is the code length, and the error-correcting power is governed by the distance between code words. The $n$-vs-Rest codes ($n$vR: the columns are all splits of $n$ classes versus the rest) include 1-vs-Rest, which has length $M$ but distance only $2$, and 2vR, whose distance $2(M-2)$ costs $\binom M2$ columns. The Hadamard code reaches distance $M/2$ with only $M-1$ columns, meeting the Plotkin bound \citep{plotkin1960binary}, and is used in ECOC practice \citep{Yin2005-nu,verma2019error,alvarez2026walsh}. Which decoder may be used with such efficient codes?

\subsection{Contributions}

Our analysis rests on one observation. In comparing classes $m$ and $m'$, only the columns on which their code words differ matter, and each casts a \textbf{vote} for $m$ or $m'$. The sum decoder decides by the sample mean of these votes, the product decoder by an M-estimator through $\artanh$. For every code, each vote is the MAP margin $\pi_m-\pi_{m'}$ plus a bias determined by the background mass alone (Section~\ref{sec:votes}). We obtain the following results.

First, the main theorem: the product decoder is MAP-consistent for all $\vpi$ in the interior of the simplex if and only if every class pair has pair structure, and if and only if the code is a union of $n$vR codes (Theorem~\ref{th:product-pair}). Pair structure means that the columns on which two classes differ are matched one to one with the columns obtained by swapping the two classes (or their complements). Prior work \citep{kumoi2021IPSJ} represented each column by the pair of class sets on its two sides and defined this correspondence as a symmetry condition. It showed that the condition suffices for product consistency and proved consistency for the $n$vR family. The main theorem strengthens this to a necessary and sufficient condition and identifies the codes that satisfy it: pair structure for every class pair makes the code invariant under every permutation of the classes, and the invariant codes are exactly the unions of $n$vR codes. For comparison we restate the known condition for the sum decoder, equidistance \citep{james1998error}, in the same terms (Proposition~\ref{prop:sum-equidistant}): the sum decoder needs only the first moment of the votes to balance, the product decoder the whole vote distribution. Every product-consistent code is therefore sum-consistent (Corollary~\ref{cor:product-implies-sum}), but not conversely.

Second, Hadamard codes ($M\geq8$) do not have a single paired column because their columns are orthogonal (Proposition~\ref{prop:hadamard-no-pair}), so for every type of Hadamard matrix the product decoder can disagree with MAP on every class pair. For the Sylvester type, the reversal margin can be at least a constant $\bar\delta\approx0.018175$ independent of $M$ (Theorem~\ref{th:worst-case}). To leading order the reversal margin is governed by the skewness of the vote sample (Eq.~\eqref{eq:beta-star-expansion}), and under typical posterior distributions it roughly decays with $M$ (Experiment~B). The efficiency of the Hadamard code and product consistency are thus incompatible.

Third, experiments on six real datasets examine which decoder is better under estimation error (Section~\ref{sec:experiments}). The product decision, a signed sum of log-odds, is invariant to a temperature common to all columns, and with calibrated classifiers the product decoder is slightly more accurate. When the calibration error differs across columns, the sum decoder, which uses bounded votes, is better.

All theorems, lemmas, propositions, and corollaries in Sections~\ref{sec:votes}--\ref{sec:hadamard}, including the proofs in the appendix, have been machine-checked in Lean~4 with Mathlib (\leanwhere).

\subsection{Relation to prior work}

\textbf{Bayes optimality of $L_1$ decoding (James and Hastie).} \citet{james1998error} (see also \citealp{james1997error}) showed that $L_1$ decoding, i.e., the sum decoder, is Bayes optimal---it returns the Bayes class whenever the binary classifiers are Bayes classifiers---if and only if the code is equidistant. Proposition~\ref{prop:sum-equidistant} restates their theorem for the order of every class pair and in the weak form. We include its short proof because the vote decomposition used there is the starting point for the product decoder.

\textbf{Robustness to estimation error (Kittler et al.).} \citet{kittler1998combining} showed that the product rule is Bayes optimal for conditionally independent representations and that the sum rule approximates it when the posteriors are close to the priors, and argued that the sum rule is robust to estimation error. Their question is which combination rule is robust to noisy estimates; ours is whether each decoder, as a function of the code $C$, reproduces MAP given the true probabilities. The two are complementary. In our real-data experiments, with calibrated classifiers the product decoder is sometimes the one that is correct when the two decoders disagree (Section~\ref{sec:experiments}).

\textbf{Asymptotic surrogate consistency (Ramaswamy et al.).} \citet{ramaswamy2014consistency} give conditions on the code matrix under which minimizing a binary surrogate loss recovers the Bayes classifier in the infinite-data limit. There the decoder is designed for the code (the inverse link of a proper composite loss followed by a linear decoding), and consistency is decided by a column-space condition on the code matrix. We go the other way: the decoder is fixed to the plain product (or sum) rule, and we characterize the codes on which it is consistent. Our study of a fixed code, noiseless probabilities, and plain decoders is thus complementary.

\textbf{Symmetry condition for the product decoder \citep{kumoi2021IPSJ}.} Prior work showed, through a symmetry condition on class pairs, that the product decoder is MAP-consistent for $n$vR codes (including 1-vs-Rest, 2vR, half-vs-half) and the Exhaustive code. We add (i)~the necessity of this symmetry condition (equivalent to pair structure) for the product decoder and the identification of all codes that satisfy it (Theorem~\ref{th:product-pair}); (ii)~the inclusion between the two decoders (Corollary~\ref{cor:product-implies-sum}); and (iii)~the absence of pair structure in Hadamard codes ($M\geq8$), the resulting inconsistency of the product decoder, and the size of its reversal margin (Section~\ref{sec:hadamard}). The Hadamard code with $M=4$ coincides with the 2vR code up to column complements and is the only Hadamard code with pair structure (Remark~\ref{rem:had4}).

\citet{passerini2004new} proposed a probability-based (product-like) decoder but gave no MAP-consistency condition. The loss-based decoding of \citet{allwein2000reducing} has the same form as the product decoder, but their theorems bound the training and generalization errors. \citet{tax2000combining} showed that sum and product coincide in the special case of two classes and two classifiers. Viewing sum and product as linear and logarithmic opinion pools \citep{degroot1991optimal,genest1986combining} places both among aggregation rules.

Section~\ref{sec:prelim} defines the notation and the decoders, Section~\ref{sec:votes} decomposes the votes, Section~\ref{sec:sum} restates the condition for the sum decoder, and Section~\ref{sec:product} proves the main theorem for the product decoder. Section~\ref{sec:hadamard} treats Hadamard codes, Section~\ref{sec:experiments} presents experiments, Section~\ref{sec:discussion} discusses, and Section~\ref{sec:conclusion} concludes. Details are in the appendix.

\section{Preliminaries}\label{sec:prelim}

\subsection{Decoders and MAP-consistency}

Let $C\in\{0,1\}^{M\times n}$ be the code matrix for $M$ classes and code length $n$ (row $m$ is the code word $\bm c_m$), $d(m,m')$ the Hamming distance between code words, and $w_m=\sum_jc_{mj}$ the row weight. Given binary-classifier probabilities $\bm p\in[0,1]^n$, the two decoding scores of class $m$ are
\begin{align}
  \text{Sum decoder:}\quad & V_m = \sum_j \bigl[c_{mj}\,p_j + (1-c_{mj})(1-p_j)\bigr],\label{eq:sum-score}\\
  \text{Product decoder (log-likelihood):}\quad & L_m = \sum_j \bigl[c_{mj}\log p_j + (1-c_{mj})\log(1-p_j)\bigr],\label{eq:product-score}
\end{align}
with predictions $\hat m=\argmax_m V_m$ and $\hat m=\argmax_m L_m$. The sum decoder is equivalent to $L_1$ decoding onto the code words in $\{0,1\}^n$ \citep{dietterich1995solving}, and the product decoder sums the Bernoulli log-likelihoods of the columns, i.e., is a logarithmic pool \citep{genest1986combining}.

We assume that no column of the code matrix is all-$0$ or all-$1$ and that all code words are distinct. Since the product score requires $p_j\in(0,1)$, statements about the product decoder take $\vpi$ in the relative interior of the simplex (all entries positive), where these assumptions give $0<p_j<1$ for every column. The sum decoder is treated on all of $\Delta^{M-1}$, including the boundary. Complementing column $j$ ($0\leftrightarrow1$) maps $p_j\to1-p_j$ and leaves the $j$th term of $V_m$ and $L_m$ unchanged: a complemented column is the same classifier for both decoders. We therefore assume that a code contains no repeated column and no pair of mutually complemented columns (repeating every column equally often leaves both decoders unchanged, and repeating only some amounts to weighting columns).

The MAP class is $\hat m^{\rm MAP}=\argmax_m\pi_m$. A decoder $D$ ($V$ or $L$) is \textbf{MAP-consistent} if, for every $\vpi$ in the range considered,
\begin{equation}\label{eq:map-consistency}
  \operatorname{sign}(D_m-D_{m'}) = \operatorname{sign}(\pi_m-\pi_{m'})\qquad(\forall m\neq m'),
\end{equation}
i.e., the preference order of the decoder agrees with the posterior order on every class pair; at a tie $\pi_m=\pi_{m'}$ we require $D_m=D_{m'}$. For the weak form, which imposes only $\pi_m>\pi_{m'}\Rightarrow D_m>D_{m'}$, Proposition~\ref{prop:sum-equidistant} and Theorem~\ref{th:product-pair} give the same conditions.

Here \textbf{noiseless} refers to the assumption that the binary classifiers output the true posteriors without error ($\bm p=C^\top\vpi$), the same term as in prior work \citep{kumoi2023IJNS}. Real classifiers have $\hat p_j\neq p_j$, introducing estimation error, which we treat in the real-data experiment of Section~\ref{sec:experiments} (Experiment~C).

\subsection{Equidistant codes and the code families studied}

\begin{definition}[Equidistant code]\label{def:equidistant}
A code matrix $C$ is \textbf{equidistant} if every pair of distinct classes has the same Hamming distance $d=d(C)$.
\end{definition}

Table~\ref{tab:code-families} lists the code families studied and our conclusions. An $n$vR code has as columns all partitions of $n$ classes versus the rest, including 1-vs-Rest ($n=1$), 2vR ($n=2$; two classes versus the rest, not one-vs-one), and half-vs-half ($n=M/2$). The normalized Hadamard code is the $M\times(M-1)$ code obtained from a normalized Hadamard matrix of order $M$ by deleting the constant column and mapping $\pm1$ to $\{1,0\}$. Class 0 has the all-ones code word, the other classes have weight $M/2-1$, and every class pair is at distance $M/2$ \citep{plotkin1960binary}. With a Sylvester matrix ($M$ a power of two), its code words are those of the simplex code with every bit complemented.

\begin{table}[tb]
  \centering
  \caption{Code families studied and their MAP-consistency (noiseless setting). Sum/Product: whether the sum/product decoder is MAP-consistent.}
  \label{tab:code-families}
  \small
  \begin{tabular}{llllll}
    \toprule
    Name & Shape & Equidistant & Pair structure & Sum & Product \\
    \midrule
    1-vs-Rest & $(M,M)$ & Yes, $d=2$ & Yes & Yes & Yes \\
    2vR & $(M,\binom{M}{2})$ & Yes, $d=2(M-2)$ & Yes & Yes & Yes \\
    Half-vs-Half & $(M,\binom{M}{M/2}/2)$ & Yes, $d=\binom{M-2}{M/2-1}$ & Yes & Yes & Yes \\
    Hadamard ($M=4$) & $(4,3)$ & Yes, $d=2$ & Yes & Yes & Yes \\
    Hadamard ($M\geq8$) & $(M,M-1)$ & Yes, $d=M/2$ & No & Yes & No \\
    logM & $(M,\lceil\log_2M\rceil)$ & No & No & No & No \\
    \bottomrule
  \end{tabular}
\end{table}

\begin{remark}[Hadamard code with $M=4$]\label{rem:had4}
The normalized Hadamard code with $M=4$ has three columns, each placing exactly two classes on one side, so its column set coincides, up to complements, with that of the 2vR code ($=$ half-vs-half($M=4$)). For $M\geq8$ it coincides with no $n$vR code. As shown in Section~\ref{sec:hadamard}, the presence of pair structure also splits at this boundary between $M=4$ and $M\geq8$.
\end{remark}

\section{Class-Pair Comparisons as Samples of Votes}\label{sec:votes}

We express the class-pair comparisons of both decoders through one object, the sample of votes cast by the columns on which the two classes differ; the results hold for every code. Fix a class pair $(m,m')$ and write $T=T_{mm'}$ for the set of columns on which $\bm c_m$ and $\bm c_{m'}$ differ, and $s=|T|=d(m,m')$. Outside $T$ we have $c_{mj}=c_{m'j}$, so only the columns in $T$ contribute to $V_m-V_{m'}$ and $L_m-L_{m'}$.

\begin{definition}[Support, vote, agreement sign]\label{def:vote}
For a column $j\in T$, the total posterior of the classes with the same bit as class $m$, $q_j:=\sum_{k:\,c_{kj}=c_{mj}}\pi_k$, is the \textbf{support} for $m$ in column $j$, and $z_j:=2q_j-1\in[-1,1]$ is the \textbf{vote} of column $j$. The vote is positive if it leans toward $m$ and negative if toward $m'$. For a background class $k\notin\{m,m'\}$, the \textbf{agreement sign} is $\tau_{kj}:=+1$ if $c_{kj}=c_{mj}$ and $-1$ otherwise, and $\bm\tau_j:=(\tau_{kj})_{k\notin\{m,m'\}}\in\{\pm1\}^{M-2}$ is the \textbf{agreement pattern} of column $j$.
\end{definition}

If $c_{mj}=1$ then $q_j=p_j$, and if $c_{mj}=0$ then $q_j=1-p_j$. Using the odd symmetry $\ell(1-p)=-\ell(p)$ of the log-odds $\ell(p)=\log\frac{p}{1-p}=2\artanh(2p-1)$, we obtain
\begin{equation}\label{eq:pair-diff}
  V_m-V_{m'}=\sum_{j\in T}z_j,\qquad L_m-L_{m'}=\sum_{j\in T}\ell(q_j)=2\sum_{j\in T}\artanh(z_j).
\end{equation}
That is, \textbf{the sum decoder decides a class pair by the sign of the sample mean of the votes, and the product decoder by the sign of the sum of the votes passed through $\artanh$}. Since $\artanh$ is strictly increasing, the latter is equivalent to taking the sign of the root $\mu$ of the location M-estimating equation $\sum_j\artanh(z_j-\mu)=0$ with influence function $\artanh$. The slope of $\artanh$ diverges as $z_j\to\pm1$, so the product decoder weights confident columns (votes near $\pm1$) heavily, the opposite of a Huber-type M-estimator, which suppresses outliers.

\begin{lemma}[Vote decomposition]\label{lem:vote-decomp}
Let $\beta:=\pi_m-\pi_{m'}$. For every $j\in T$,
\begin{equation}\label{eq:vote-decomp}
  z_j=\beta+a_j,\qquad a_j:=\sum_{k\notin\{m,m'\}}\pi_k\,\tau_{kj}.
\end{equation}
The bias $a_j$ is determined by the background posteriors $\bm b:=(\pi_k)_{k\notin\{m,m'\}}$ alone and does not involve $\pi_m,\pi_{m'}$.
\end{lemma}

\begin{proof}
For $j\in T$ we have $c_{mj}\neq c_{m'j}$, so class $m$ belongs to $q_j$ and class $m'$ to $1-q_j$. Hence $z_j=q_j-(1-q_j)=\pi_m-\pi_{m'}+\sum_{k\notin\{m,m'\}}\pi_k\tau_{kj}$.
\end{proof}

\begin{figure}[tb]
  \centering
  \includegraphics[width=\linewidth]{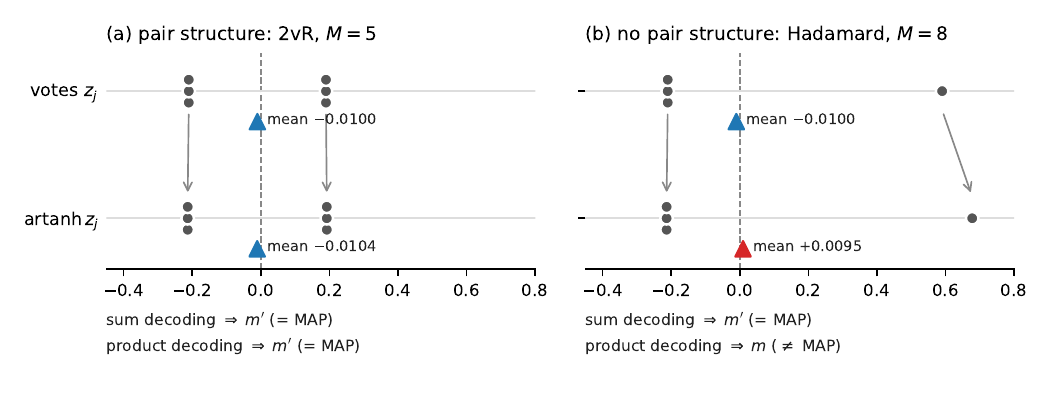}
  \caption{Vote sample of one class pair, with the same masses in both panels ($\pi_m=0.195$, $\pi_{m'}=0.205$, $0.2$ on each of three background classes; MAP picks $m'$). The sum decoder decides by the sign of the mean of the votes $z_j$, the product decoder by the sign of the mean of $\artanh z_j$. (a)~With pair structure the sample is symmetric and the two decoders agree. (b)~For the Hadamard code with $M=8$ the sample is skewed; $\artanh$ stretches the large vote, and only the product decoder reverses. The masses in (b) are the construction of Theorem~\ref{th:worst-case}, with class indices translated in $\mathbb F_2^t$.}
  \label{fig:vote-concept}
\end{figure}

By Lemma~\ref{lem:vote-decomp}, fixing the background mass $\bm b$ and moving mass only between $m$ and $m'$ translates the vote sample $\{z_j\}_{j\in T}$ by $\beta$ without changing its shape. We work on this one-dimensional family. Since $\pi_m+\pi_{m'}=1-\sum_kb_k=:r$, $\beta$ ranges over $[-r,r]$ on it ($(-r,r)$ for the product decoder). Figure~\ref{fig:vote-concept} gives an example: with pair structure (Section~\ref{sec:product}) the sample is symmetric and the two decoders agree; without it, $\artanh$ stretches the large votes of a skewed sample and only the product decoder can reverse the MAP order.

The sum of the biases is expressed by code distances.

\begin{lemma}[Distance identity]\label{lem:distance-identity}
For every background class $k\notin\{m,m'\}$, $\sum_{j\in T}\tau_{kj}=d(k,m')-d(k,m)$. Hence
\begin{equation}\label{eq:distance-identity}
  V_m-V_{m'}=d(m,m')\,(\pi_m-\pi_{m'})+\sum_{k\notin\{m,m'\}}\pi_k\bigl[d(k,m')-d(k,m)\bigr].
\end{equation}
\end{lemma}

\begin{proof}
For $j\in T$, the bit of $k$ agrees with exactly one of $m$ and $m'$. Thus $\sum_{j\in T}\tau_{kj}$ is the number of columns in $T$ on which $k$ agrees with $m$ minus the number on which it agrees with $m'$, i.e., $d_T(k,m')-d_T(k,m)$ for the distance $d_T$ restricted to $T$. Outside $T$, $\bm c_m$ and $\bm c_{m'}$ agree, so the contributions outside $T$ are equal in both distances and the difference is $d(k,m')-d(k,m)$. Eq.~\eqref{eq:distance-identity} follows by substituting into Eqs.~\eqref{eq:pair-diff} and~\eqref{eq:vote-decomp}.
\end{proof}

The second term of Eq.~\eqref{eq:distance-identity} says that \emph{each background class $k$ pulls the sum decision toward the class closer to itself}: if $k$ is closer to $m$ ($d(k,m)<d(k,m')$), the decision shifts toward $m$.

\section{Sum Decoder: Equidistance Is Necessary and Sufficient}\label{sec:sum}

The condition for the sum decoder is known \citep{james1998error}. We restate it in the language of votes, which sets up the comparison with the product decoder.

\begin{proposition}[Sum decoder; \citealp{james1998error}]\label{prop:sum-equidistant}
Let $M\geq3$. The sum decoder is MAP-consistent for all $\vpi\in\Delta^{M-1}$ if and only if the code is equidistant, in which case $V_m-V_{m'}=d(C)\,(\pi_m-\pi_{m'})$. The equivalence also holds for the weak form of consistency that does not require ties.
\end{proposition}

\begin{proof}
(Sufficiency) If the code is equidistant, the second term of Eq.~\eqref{eq:distance-identity} vanishes and $V_m-V_{m'}=d(C)(\pi_m-\pi_{m'})$. Since $d(C)\geq1$, the signs agree.

(Necessity) Take distinct $k,m,m'$, and for $t\in(0,1)$ and $\varepsilon$ with $|\varepsilon|<(1-t)/2$ set $\pi_m=\frac{1-t}{2}+\varepsilon$, $\pi_{m'}=\frac{1-t}{2}-\varepsilon$, $\pi_k=t$, and all others $0$. By Eq.~\eqref{eq:distance-identity}, $V_m-V_{m'}=2\varepsilon\,d(m,m')+t\,[d(k,m')-d(k,m)]$. Even under weak consistency this must be positive for $\varepsilon>0$ and negative for $\varepsilon<0$, so letting $\varepsilon\to0$ gives $t\,[d(k,m')-d(k,m)]=0$, i.e., $d(k,m)=d(k,m')$. Since this holds for any three distinct classes, two class pairs sharing a class have equal distances. Any two class pairs $\{a,b\},\{c,e\}$ are connected by a chain of pairs each sharing a class with the next, such as $\{a,b\}\to\{b,c\}\to\{c,e\}$ (a shorter chain suffices when, e.g., $b=c$), so all class pairs have equal distances.
\end{proof}

For $M=2$, $V_1-V_2=d(1,2)(\pi_1-\pi_2)$ for distinct code words, so the sum decoder is always MAP-consistent.

Equidistance can also be written as a Gram-matrix condition.

\begin{corollary}[Gram-matrix form]\label{cor:equidistant-equiv}
A code is equidistant if and only if there exist $\lambda>0$ and $\bm\omega\in\mathbb R^M$ such that $G=CC^\top=\lambda I+\bm\omega\bm1^\top+\bm1\bm\omega^\top$, in which case $d(C)=2\lambda$. In the equal-weight case ($w_m$ constant), $\bm\omega$ is a constant vector and the condition reduces to $G=\lambda I+\theta\bm1\bm1^\top$.
\end{corollary}

\begin{proof}
($\Leftarrow$) $d(m,m')=G_{mm}+G_{m'm'}-2G_{mm'}=2\lambda$. ($\Rightarrow$) Setting $\lambda:=d/2$ and $\omega_m:=w_m/2-d/4$ gives $G_{mm}=w_m=\lambda+2\omega_m$ and $G_{mm'}=(w_m+w_{m'}-d)/2=\omega_m+\omega_{m'}$.
\end{proof}

\subsection{Application to code families}

All code families in Table~\ref{tab:code-families} except logM are equidistant. The Hadamard code has unequal row weights (the all-ones row $M-1$, the others $M/2-1$) but is equidistant, so the sum decoder is MAP-consistent. logM violates equidistance, and the sum decoder indeed disagrees with MAP (Section~\ref{sec:experiments}).

The order of a Hadamard matrix is a multiple of four (apart from $1,2$), but equidistance is preserved under row deletion, since deleting rows but keeping columns preserves the distances among the remaining rows.

\begin{corollary}[Equidistant codes by row deletion]\label{cor:subcode}
Deleting two rows from the Hadamard code with $M=12$ yields an equidistant code with $M=10$. Deleting two rows including the all-ones row (weight $M-1$) leaves ten rows forming an equal-weight equidistant code; keeping the all-ones row and deleting two other rows gives an unequal-weight equidistant code. In both cases the sum decoder is MAP-consistent.
\end{corollary}
\noindent Numerically, no noiseless sum-decoder disagreement was observed for either deletion (Appendix~\ref{app:protocol}). In general, for $M$ not a multiple of four, deleting rows from the Hadamard code of order $M'=4\lceil M/4\rceil$ (as done in practice, e.g., by \citealp{verma2019error}) gives an equidistant code for $M$ classes with $M'-1\leq M+2$ columns and distance $M'/2\geq M/2$: the cost of the missing orders is at most three extra columns.

\begin{remark}[Sum $\equiv$ product $\equiv$ bitwise thresholding on complete codes]\label{rem:complete-code}
When $M$ is a power of two, the logM code has every word of $\{0,1\}^{\log_2M}$ as a code word (a complete code). Each column term can then be maximized independently, so the sum decoder, the product decoder, and bitwise thresholding coincide identically and the two decoders never differ, while their disagreement with MAP equals the gap between the bitwise marginal MAP and the joint MAP. The disagreement of logM is thus not a matter of sum versus product but a structural consequence of cutting the code length to $\log_2M$.
\end{remark}

\section{Product Decoder: The Main Theorem}\label{sec:product}

The sum decoder looks only at the mean of the votes; through $\artanh$, the product decoder also sees the shape of the vote sample. We characterize product reversals on the one-dimensional family with fixed background (Proposition~\ref{prop:pair-inversion}) and then give the necessary and sufficient condition over all $\vpi$ (Theorem~\ref{th:product-pair}). Throughout this section $\vpi$ lies in the relative interior of the simplex.

\subsection{Reversal on a single class pair}

Fix a class pair $(m,m')$ and a background $\bm b$ (all entries positive, $r=1-\sum_kb_k>0$), and let
\begin{equation}\label{eq:psi-def}
  \Psi_{\bm b}(\beta):=\sum_{j\in T}\artanh(\beta+a_j),
\end{equation}
where $a_j$ is the bias of Lemma~\ref{lem:vote-decomp}. By Eqs.~\eqref{eq:pair-diff} and~\eqref{eq:vote-decomp}, $L_m-L_{m'}=2\Psi_{\bm b}(\beta)$ on the family.

\begin{proposition}[Reversal on a single class pair]\label{prop:pair-inversion}
$\Psi_{\bm b}$ is strictly increasing on $\beta\in(-r,r)$.
\begin{enumerate}
  \item If $\Psi_{\bm b}(0)=0$, then $\operatorname{sign}(L_m-L_{m'})=\operatorname{sign}\beta$ for every $\beta$ on the family (no reversal occurs).
  \item If $\Psi_{\bm b}(0)>0$, then $\Psi_{\bm b}$ has a unique root $\beta^*<0$, and the product decoder reverses the MAP order if and only if $\beta\in(\max(\beta^*,-r),\,0)$. This interval is nonempty. The case $\Psi_{\bm b}(0)<0$ follows by swapping $m\leftrightarrow m'$.
\end{enumerate}
\end{proposition}

Proposition~\ref{prop:pair-inversion} (proved in Appendix~\ref{app:proofs}) shows that reversal on a class pair is determined solely by \emph{whether the bias sample $\{a_j\}$ is balanced at zero in the $\artanh$ sense} (whether $\sum_j\artanh(a_j)=0$). Even biases with mean zero (as for every equidistant code) need not be balanced, since $\artanh$ is nonlinear.

\subsection{Pair structure and the main theorem}

\begin{definition}[Pair structure]\label{def:pair-structure}
For a class pair $(m,m')$, columns $j,j'\in T$ \textbf{are paired} if column $j'$ equals the column obtained from column $j$ by swapping its $m$th and $m'$th entries, or the complement of that column. The code has \textbf{pair structure} for $(m,m')$ if $T$ can be partitioned exactly into paired columns.
\end{definition}

Complements are allowed because a complemented column represents the same classifier (Section~\ref{sec:prelim}). The symmetry condition of prior work \citep{kumoi2021IPSJ} is defined on the family of class sets on both sides of each column, so it does not distinguish complements either; since for $M\geq3$ the swap has no fixed point on $T$, pair structure for all class pairs is equivalent to that condition.

\begin{lemma}[Pairs and agreement patterns]\label{lem:pair-pattern}
Let $M\geq3$. Columns $j,j'\in T$ are paired if and only if $\bm\tau_{j'}=-\bm\tau_j$. Hence the class pair $(m,m')$ has pair structure if and only if the multiset of agreement patterns $\{\bm\tau_j\}_{j\in T}$ is invariant under the sign flip $\bm\tau\mapsto-\bm\tau$.
\end{lemma}

\begin{proof}
In the $\pm1$ representation $h_{kj}:=2c_{kj}-1$, we have $\tau_{kj}=h_{mj}h_{kj}$ and $h_{m'j}=-h_{mj}$ ($j\in T$). The swapped column of column $j$ has entry $-h_{mj}$ at $m$, $h_{mj}$ at $m'$, and background entries $h_{kj}$, so its agreement pattern is $(-h_{mj})h_{kj}=-\tau_{kj}$. Complementing all entries flips the signs of both $h_{mj'}$ and $h_{kj'}$, leaving the agreement pattern unchanged. Hence paired columns satisfy $\bm\tau_{j'}=-\bm\tau_j$. Conversely, suppose $\bm\tau_{j'}=-\bm\tau_j$. If $h_{mj'}=-h_{mj}$, the background entries are $h_{kj'}=h_{kj}$ and column $j'$ equals the swapped column. If $h_{mj'}=h_{mj}$, then $h_{kj'}=-h_{kj}$ and column $j'$ equals the complement of the swapped column. For the second statement, $\bm\tau_j\neq-\bm\tau_j$ when $M\geq3$, so if the multiset is invariant under sign flip, each occurrence of a pattern $\bm v$ can be matched one to one with an occurrence of $-\bm v$, and conversely.
\end{proof}

We read each column $j$ as the split $\{A_j,A_j^c\}$ of the classes, where $A_j=\{k:c_{kj}=1\}$.

\begin{definition}[Union of $n$vR codes]\label{def:nvr-union}
A code is a \textbf{union of $n$vR codes} if its columns, read as splits, are exactly the splits of the $n$vR codes for $n$ in some nonempty set $N\subseteq\{1,\ldots,\lfloor M/2\rfloor\}$; equivalently, together with every column, every split with the same side sizes is a column.
\end{definition}

Examples are each $n$vR code ($N=\{n\}$), the Exhaustive code ($N=\{1,\ldots,\lfloor M/2\rfloor\}$), and 1-vs-Rest together with 2vR ($N=\{1,2\}$).

\begin{theorem}[Product decoder; main theorem]\label{th:product-pair}
Let $M\geq3$, take $\vpi$ in the relative interior of the simplex, and let the code satisfy the conventions of Section~\ref{sec:prelim} (distinct code words, no repeated or complemented columns). The following are equivalent: (1)~the product decoder is MAP-consistent; (2)~the product decoder is MAP-consistent in the weak form ($\pi_m>\pi_{m'}\Rightarrow L_m>L_{m'}$); (3)~every class pair has pair structure; (4)~the code is a union of $n$vR codes.
\end{theorem}

\begin{proof}
(3)$\Rightarrow$(1): By Lemma~\ref{lem:pair-pattern}, paired $j,j'$ satisfy $a_{j'}=\langle\bm b,\bm\tau_{j'}\rangle=-a_j$. Hence $\Psi_{\bm b}(0)=\sum_{j\in T}\artanh(a_j)=0$ for every background $\bm b$, and by Proposition~\ref{prop:pair-inversion}(1) the signs agree for all pairs and all $\vpi$. (1)$\Rightarrow$(2): Immediate.

(2)$\Rightarrow$(3): We prove the contrapositive. If a class pair $(m,m')$ lacks pair structure, it suffices to show $\Psi_{\bm b}(0)\neq0$ for some background $\bm b$; then by Proposition~\ref{prop:pair-inversion}(2) the product decoder strictly reverses the MAP order on a nonempty interval of $\beta$, contradicting weak consistency. In outline: Expanding $\Psi_{\bm b}(0)=\sum_{j\in T}\artanh\langle\bm b,\bm\tau_j\rangle$ as a power series in $\bm b$, the term of degree $\kappa$ is $\frac1\kappa\sum_{j\in T}\langle\bm b,\bm\tau_j\rangle^\kappa$ ($\kappa$ odd). If $\Psi_{\bm b}(0)=0$ for every background, all these terms vanish, and all odd moments of the distribution of agreement patterns are $0$. For a distribution on a finite set, the vanishing of all odd moments is equivalent to symmetry under sign flip, which means pair structure (Lemma~\ref{lem:pair-pattern}). Appendix~\ref{app:proof-product} makes this rigorous by restricting $\bm b$ to a single ray $\bm b=t\bm v$, which needs only a one-variable power series and a Vandermonde matrix.

(3)$\Leftrightarrow$(4): A column with $c_{mj}=c_{m'j}$ is unchanged by swapping $m$ and $m'$. Hence pair structure for $(m,m')$ says exactly that the set of columns, read as splits, is invariant under the transposition of $m$ and $m'$. Pair structure for every class pair is thus invariance under every transposition and, since transpositions generate the symmetric group, under every permutation of the classes. The orbit of a split $\{A,A^c\}$ under all permutations consists of all splits with the same side sizes, i.e., the columns of the $n$vR code with $n=\min(|A|,M-|A|)$. A set of splits invariant under all permutations is therefore a union of such orbits, which is (4); conversely, a union of $n$vR codes is invariant under every permutation.
\end{proof}

Without the convention on repeated columns, (4) reads: the multiplicity of a split among the columns depends only on its side sizes.

\begin{corollary}[Product consistency $\Rightarrow$ sum consistency]\label{cor:product-implies-sum}
A code on which the product decoder is MAP-consistent is equidistant, and hence the sum decoder is also MAP-consistent. The converse fails (Hadamard codes, $M\geq8$; Section~\ref{sec:hadamard}).
\end{corollary}

\begin{proof}
Under pair structure, the pairs $\bm\tau_{j'}=-\bm\tau_j$ cancel in the sum, so $\sum_{j\in T}\tau_{kj}=0$ for every background class $k$. By Lemma~\ref{lem:distance-identity}, $d(k,m)=d(k,m')$, and the argument in the necessity part of Proposition~\ref{prop:sum-equidistant} shows that the code is equidistant.
\end{proof}

Proposition~\ref{prop:sum-equidistant} and Theorem~\ref{th:product-pair} are requirements of different strength on the vote biases $\{a_j\}$. The sum decoder requires that the \emph{sum} of the agreement signs of each background class vanish (balance of the first moment); this is equidistance. The product decoder requires that the \emph{distribution} of agreement patterns itself be symmetric under sign flip (balance of all odd moments); this is pair structure. First-order balance alone is not enough for the product decoder, and through condition (4) the symmetry of the vote distribution for every class pair becomes a symmetry of the whole code under all permutations of the classes.

\subsection{Application to code families}

\begin{corollary}[Code families with pair structure]\label{cor:families}
The $n$vR codes (including 1-vs-Rest, 2vR, and half-vs-half), the Exhaustive code, their unions, and the Hadamard code with $M=4$ have pair structure for every class pair, so both decoders are MAP-consistent on them.
\end{corollary}

\begin{proof}
Each is a union of $n$vR codes: the Exhaustive code is the union over all $n$, and the columns of the Hadamard code with $M=4$ are the three splits of the four classes into two pairs, i.e., half-vs-half with $M=4$ (Remark~\ref{rem:had4}). Condition (4) of Theorem~\ref{th:product-pair} applies. For pairs of the Hadamard code with $M=4$ containing class 0 (the all-ones row), both columns of $T$ lie on the side of class 0, so the pair is given not by the swap but by the complement of the swap.
\end{proof}

As in the last case, some columns pair only with the complement of the swapped column when read as 0/1 vectors; representing each column by the class sets on its two sides removes the distinction (the Hadamard code with $M=4$ is the same matrix as the 2vR example ($M=4$) of prior work).

Pair structure is fragile. Deleting one column from 2vR breaks its pair, and pair structure is lost for the class pairs that differ on that column. This is consistent with the jump in the product disagreement rate (from 0\% to a few \%) observed in Experiment~A. Row deletion, on the other hand, preserves pair structure (the swap and complement relations still hold on the remaining rows).

\begin{remark}[The cost of product consistency]\label{rem:cost}
By condition (4), a product-consistent code contains an entire $n$vR block. For $M\geq8$, every product-consistent code other than 1-vs-Rest therefore has at least $\binom M2$ columns ($\binom Mn\geq\binom M2$ for $2\leq n\leq M-2$, and $\binom{M}{M/2}/2\geq\binom M2$ for $M\geq8$), while 1-vs-Rest has distance only $2$. The Hadamard code reaches distance $M/2$ with $M-1$ columns, fewer than 2vR by a factor of about $M/2$, but has no pair structure (Section~\ref{sec:hadamard}).
\end{remark}

\section{Hadamard Codes: Absence of Pair Structure and the Reversal Margin}\label{sec:hadamard}

Hadamard codes are equidistant, so the sum decoder is consistent. For the product decoder we show that for $M\geq8$ no column has a paired column, and we evaluate how large the resulting disagreement can be.

\subsection{Absence of pair structure}

\begin{proposition}[Hadamard codes have no pair structure]\label{prop:hadamard-no-pair}
For every normalized Hadamard code with $M\geq8$ (not only the Sylvester type) and every class pair $(m,m')$, no column in $T$ has a paired column.
\end{proposition}

\begin{proof}
In the $\pm1$ representation, suppose columns $j,j'\in T$ are paired. The swapped column differs from column $j$ only in entries $m$ and $m'$, so the entrywise product $\bm h_{\cdot j}\circ\bm h_{\cdot j'}$ has exactly $2$ entries equal to $-1$, and for the complement of the swapped column exactly $M-2$. On the other hand, distinct columns of a Hadamard matrix are orthogonal, so the entrywise product sums to $0$, i.e., has exactly $M/2$ entries equal to $-1$. We have $M/2\in\{2,M-2\}$ only for $M=4$.
\end{proof}

\begin{corollary}[Product inconsistency on Hadamard codes]\label{cor:hadamard-inconsistent}
For every normalized Hadamard code with $M\geq8$, the sum decoder is MAP-consistent but the product decoder is not. Moreover, for every class pair there exists $\vpi$ on which the product decoder reverses the MAP order.
\end{corollary}

\begin{proof}
By Proposition~\ref{prop:hadamard-no-pair} no class pair has pair structure, so by the proof of (2)$\Rightarrow$(3) in Theorem~\ref{th:product-pair}, for every class pair there exists $\vpi$ at which a reversal occurs.
\end{proof}

The columns of a Hadamard code are maximally spread: any two differ on half of the rows. This enlarges the code distance and gives equidistance (first-order balance), but makes it impossible for two columns to be related by a swap (pair structure). The property that makes the Hadamard code a good error-correcting code thus rules out pair structure. Orthogonality is also why \citet{verma2019error} choose Hadamard codes, since orthogonal code words make their probability estimates admissible; for MAP-consistency of the product decoder, the orthogonality of the columns is instead the obstacle.

For the Sylvester type, some class does not appear at all in the comparison statistic.

\begin{proposition}[Invisible class]\label{prop:invisible-class}
For the Sylvester Hadamard code with $M=2^t$, index the classes by $\mathbb F_2^t$. For every class pair $(m,m')$, the class $k^*=m\oplus m'$ satisfies $c_{k^*j}=0$ identically on $T$. Hence $p_j$ on $T$ does not involve $\pi_{k^*}$, and $L_m-L_{m'}$ does not depend on $\pi_{k^*}$.
\end{proposition}

\begin{proof}
The code word of class $a\in\mathbb F_2^t$ on the nonzero columns $j\in\mathbb F_2^t\setminus\{0\}$ is $c_{aj}=(1+\chi_a(j))/2$ with $\chi_a(j)=(-1)^{\langle a,j\rangle}$ (since $\chi_0\equiv1$, $\bm c_0=\bm1$). By the character product rule $\chi_m\chi_{m'}=\chi_{m\oplus m'}$, $T=\{j:\chi_m(j)\chi_{m'}(j)=-1\}=\{j:\chi_{m\oplus m'}(j)=-1\}$, so $c_{k^*j}=0$ for $j\in T$.
\end{proof}

For a pair $(0,m)$ containing class 0, $k^*=m$: $L_0-L_m$ does not involve $\pi_m$ of the compared class itself, whereas the sum decoder gives $V_0-V_m=\frac M2(\pi_0-\pi_m)$, which involves $\pi_m$ explicitly.

\subsection{How large can the reversal margin be?}\label{sec:worst-case}

By Proposition~\ref{prop:pair-inversion}, the width of the range of $\beta$ with reversal (the reversal margin) is the root $|\beta^*|$ of $\Psi_{\bm b}$ (unless it exceeds the feasible range $r$). For equidistant codes the biases have mean $\langle a\rangle:=\frac1s\sum_{j\in T}a_j=0$ (Lemma~\ref{lem:distance-identity}). Substituting $\artanh(y)=y+y^3/3+y^5/5+\cdots$ into $\Psi_{\bm b}(\beta^*)=0$ and solving for $\beta^*$ with $\langle\cdot\rangle=\frac1s\sum_{j\in T}$, the leading terms are
\begin{equation}\label{eq:beta-star-expansion}
  \beta^* \approx -\,\frac{\tfrac13\langle a^3\rangle+\tfrac15\langle a^5\rangle+\cdots}{1+\langle a^2\rangle+\cdots}.
\end{equation}
The numerator has only odd moments (skewness) of the bias sample; the even moments (variance) enter the denominator and suppress the margin. A symmetric bias sample thus produces no reversal (pair structure), and a larger reversal margin requires a skewed sample.

One may expect the background to be diluted and the margin to shrink to $0$ as $M$ grows; under typical posteriors this indeed happens (see below). For the Sylvester type, however, concentrating the background mass on a few classes to maximize skewness keeps the margin at a constant independent of $M$.

\begin{theorem}[Dimension-independent lower bound on the worst-case reversal margin]\label{th:worst-case}
Let $\bar\delta$ be the unique positive root of
\begin{equation}\label{eq:worst-case-eq}
  3\,\artanh(0.2+\delta) = \artanh(0.6-\delta)
\end{equation}
($\bar\delta\approx0.018175$). For every Sylvester Hadamard code ($M=2^t\geq8$), every class $m\neq0$, and every $\varepsilon\in(0,\bar\delta)$, there exists $\vpi$ (in the relative interior) with $\pi_m-\pi_0=\varepsilon$ such that class $m$ is the MAP class (the largest entry of $\vpi$) but the product decoder ranks class $0$ above $m$. That is, the worst-case reversal margin is at least $\bar\delta$, and $\bar\delta$ does not depend on $M$.
\end{theorem}

The key to the proof is to place mass $1/5$ on each element of an $\mathbb F_2$-closed triple $\{b_1,b_2,b_1\oplus b_2\}$ of background classes. By the character product rule, the bias $a_j$ then degenerates to two points, $+0.6$ (proportion $1/4$) and $-0.2$ (proportion $3/4$), and the root equation of $\Psi$ becomes Eq.~\eqref{eq:worst-case-eq}, which involves neither $s$ nor $M$. The full proof is given in Appendix~\ref{app:proofs}.

Numerically, $|\beta^*|=0.018175$ agrees to machine precision for $M\in\{8,16,32,64\}$, and the same construction gives not only a class-pair reversal but also an argmax disagreement. The argmax disagreement persists while the MAP margin (the gap between the top two posteriors) is at most $0.911\,\bar\delta\approx0.0166$, independently of $M$; the product decoder then picks a class of the background triple. Below, the reversal margin $\delta^*$ measured in experiments is this MAP margin at points where the argmax disagrees.

$\bar\delta$ is a proven \emph{lower} bound, not the worst-case supremum. We ran multi-start SLSQP from class pairs whose code words have equal weight and from sampled disagreement points. For every $M\in\{8,12,16,20,28,32,48,64\}$ it found margins $\delta^*_{\rm worst}\in[0.017,0.025]$ (mean $0.0232$, coefficient of variation $0.10$), each verified with the decoders. These exceed the construction value $0.0166$ for every $M$. The worst MAP margin with argmax disagreement is thus at least $0.0166$ for the Sylvester type, and local search found none above $0.025$; its exact value is open. For non-Sylvester types the triple character identity is unavailable, so we have no proved lower-bound constant (numerically, a similar $\Theta(10^{-2})$ is observed).

\subsection{The typical margin decays with $M$}

The largest $\delta^*$ among disagreement points found by random sampling (the typical value below) roughly decays with $M$ (Dirichlet: $0.0135$ at $M=8$, $0.0066$ at $M=16$, about $0.003$ at $M=48,64$). The product $\delta^*\times M$ ranges over $0.10$--$0.18$ rather than being constant, so a pure $1/M$ law does not hold. The decay is consistent with Eq.~\eqref{eq:beta-star-expansion}: as $M$ grows, the background mass spreads over many classes and the skewness $\langle a^3\rangle$ of the bias sample shrinks. The worst-case construction lies in a corner of the simplex with mass on only five classes, which random sampling rarely hits.

Hence, \emph{the worst-case reversal margin does not vanish as $M$ grows, but the $\vpi$ at which reversals occur are confined to a thin region near ties, which shrinks with $M$ under typical distributions.}

\section{Experiments}\label{sec:experiments}

We examine three questions: A,~do the theorems hold numerically (noiseless setting)? B,~how large are the reversals on Hadamard codes? C,~which decoder is more accurate with real classifiers? A and B confirm the theory; C answers the second half of the question in the introduction (which decoder is better under estimation error). Settings are detailed in Appendix~\ref{app:protocol}.

\subsection{A: Numerical confirmation of the theorems}

\begin{figure}[!tb]
  \centering
  \begin{minipage}[t]{0.57\linewidth}
    \centering
    \includegraphics[width=\linewidth]{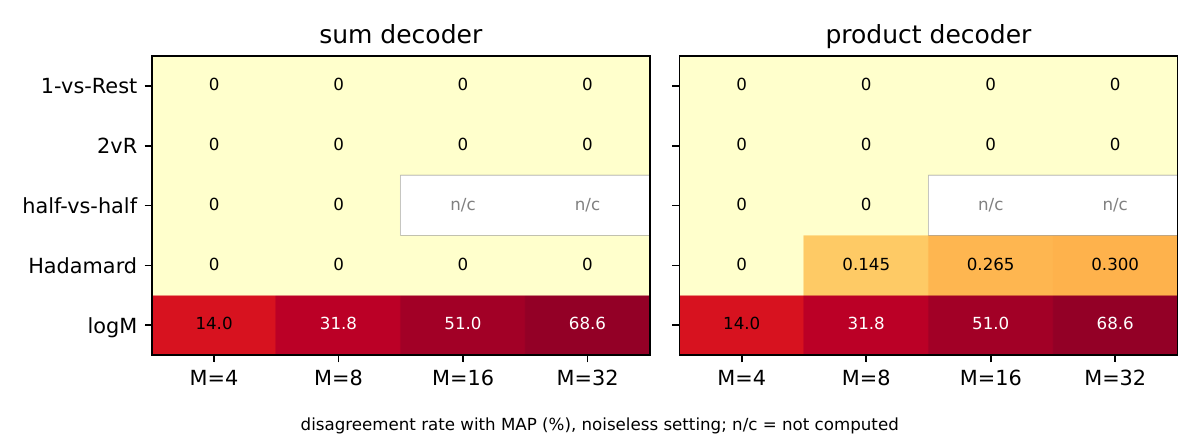}
  \caption{Disagreement rate with MAP in the noiseless setting (\%, mean over four series). Left: sum; right: product. The sum decoder is $0$ on all equidistant codes (all but logM), the product decoder only on codes with pair structure (1-vs-Rest, 2vR, half-vs-half, Hadamard with $M=4$) (Proposition~\ref{prop:sum-equidistant} and Theorem~\ref{th:product-pair}). n/c: not computed. Half-vs-half has $\binom{M}{M/2}/2$ columns (6435 at $M=16$, about $3\times10^8$ at $M=32$), so only $M\leq8$ was computed; having pair structure, its rate is $0$ by the theorem.}
  \label{fig:disagreement-heatmap}
  \end{minipage}\hfill
  \begin{minipage}[t]{0.40\linewidth}
    \centering
    \includegraphics[width=\linewidth]{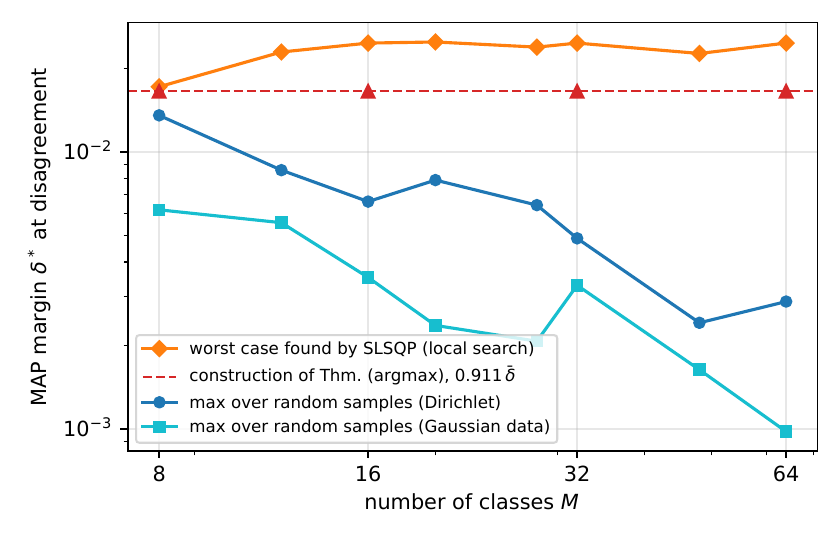}
  \caption{Dependence of the reversal margin on $M$ (Hadamard codes). Typical values (largest margin at disagreement points found by Dirichlet sampling and Gaussian data) roughly decay with $M$; the worst values found by local SLSQP search stay within $[0.017,0.025]$ regardless of $M$. Red dashed line: the largest MAP margin $0.911\,\bar\delta$ with argmax disagreement in the construction of Theorem~\ref{th:worst-case}; triangles: Sylvester-type $M$ (to which the theorem applies).}
  \label{fig:delta-M}
  \end{minipage}
\end{figure}

We sampled noiseless probabilities $\bm p=C^\top\vpi$ (four series per code and $M$: Dirichlet $\alpha\in\{0.3,1.0\}$ and two posteriors built from Gaussian data, $200{,}000$ samples each, or $100{,}000$ for codes with many columns) and measured each decoder's disagreement rate with MAP (Figure~\ref{fig:disagreement-heatmap}). The sum decoder never disagreed on the equidistant codes (all but logM). The product decoder never disagreed on 1-vs-Rest, 2vR, and half-vs-half, which have pair structure, or on the Hadamard code with $M=4$; on Hadamard codes with $M\geq8$ its rate was small but nonzero ($0.145\%$, $0.265\%$, and $0.300\%$ for $M=8,16,32$). logM disagrees heavily for both decoders ($14.0\%$ to $68.6\%$ for $M=4$ to $32$).

The experiments also show that pair structure is fragile (Section~\ref{sec:product}). Deleting one random column from 2vR raises the product disagreement rate to $4.72\%$ at $M=8$ and $3.12\%$ at $M=16$ (mean over 10 seeds). Since the deletion also breaks equidistance, the sum rate rises similarly ($5.26\%$, $4.48\%$).

To verify the main theorem pair by pair, we took all 977 class pairs of 49 codes ($n$vR-type codes; Hadamard $M\in\{4,8,12,16\}$, with $M=12$ of Paley type; logM; 24 random codes; and 10 codes obtained from $n$vR codes by row deletion) and independently (a)~tested pair structure (Lemma~\ref{lem:pair-pattern}) and (b)~searched for $\vpi$ at which the product decoder strictly reverses the MAP order (sampling near ties and local optimization). Every reversal candidate was confirmed with the original decoding formula. Reversals were found for all 773 pairs without pair structure and for none of the 204 pairs with it. Among Hadamard codes, all pairs had pair structure for $M=4$ and none for $M=8,12,16$ (consistent with Proposition~\ref{prop:hadamard-no-pair}).

\subsection{B: Size of the reversal margin}

We measured the reversal margin of Hadamard codes for $M\in\{8,12,16,20,28,32,48,64\}$ in the typical case (Dirichlet sampling, Gaussian data) and the worst case (multi-start SLSQP; Figure~\ref{fig:delta-M}). The typical value (largest $\delta^*$ at sampled disagreement points) roughly decays from $0.0135$ ($M=8$) to about $0.003$ ($M=48,64$), whereas the worst case stays within $[0.017,0.025]$ for all $M$, from 1.3 times the typical (Dirichlet) value at $M=8$ to about 9 times at $M=48,64$. Over $500{,}000$ noiseless samples for $M\in\{8,16,32\}$, the true MAP margin at disagreement points was at most $0.00845$ ($0.51$ times the construction value $0.0166$). Random posteriors rarely hit the worst-case construction (a corner of the simplex).

\subsection{C: Which decoder is more accurate with real classifiers?}

We compared the decoders under estimated probabilities on six datasets (letter, isolet, pendigits, MNIST, Fashion-MNIST, MobileNetV2 features of CIFAR-100). For random class subsets with $M\in\{8,16,24,32\}$ (as the data allow), logistic regression was trained for each column of 1-vs-Rest, 2vR, Hadamard, and random dense codes ($n=\lceil10\log_2M\rceil$). Each column was temperature-calibrated on validation data (``calibrated'' below), and calibration error was then introduced by scaling the logits (a)~of all columns by a common factor $\gamma\in\{0.5,1,2,4\}$ ($\gamma>1$: uniform overconfidence), or (b)~by column-specific factors $\gamma_j=\exp(\sigma g_j)$, $g_j\sim\mathcal N(0,1)$, $\sigma\in\{0,0.5,1\}$ (calibration error varying across columns). We compared test accuracies over 5 seeds per condition. Condition (a) and the hypotheses were fixed before the runs; (b) is exploratory, added after seeing (a). On 1-vs-Rest the decoders coincide identically (both take the argmax of the logit of column $m$), so we report the other codes.

\begin{figure}[!tb]
  \centering
  \begin{minipage}[c]{0.56\linewidth}
    \includegraphics[width=\linewidth]{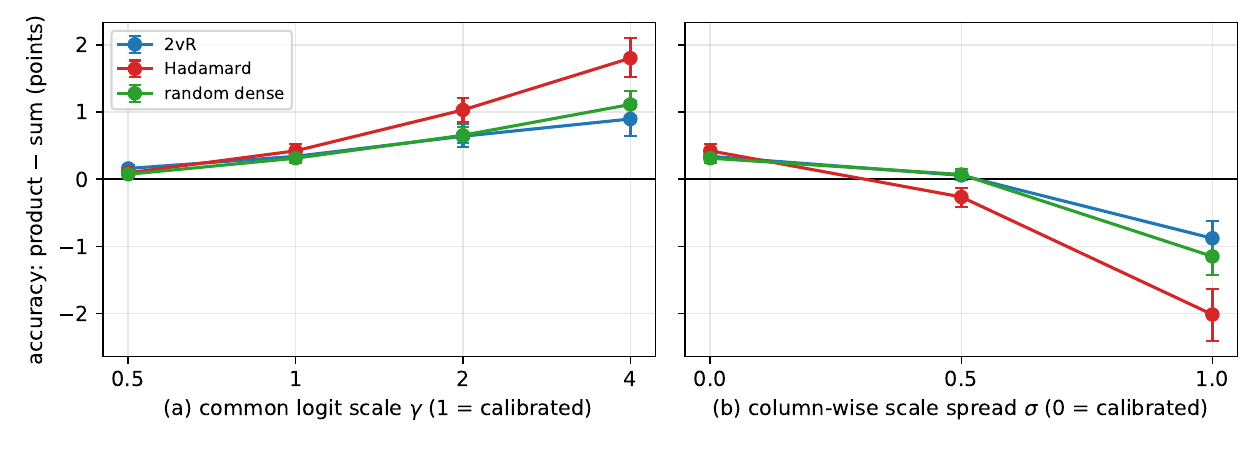}
  \end{minipage}\hfill
  \begin{minipage}[c]{0.41\linewidth}
  \caption{Accuracy difference between the product and sum decoders with real classifiers (points; positive means the product decoder is more accurate). Means and 95\% bootstrap confidence intervals over six datasets, each $M$, and 5 seeds. (a)~All logits multiplied by a common factor $\gamma$: the product decision is unchanged, while the sum decoder degrades as $\gamma$ grows. (b)~Column-specific factors $\gamma_j=\exp(\sigma g_j)$: as the spread grows, the sum decoder becomes better.}
  \label{fig:real-decoders}
  \end{minipage}
\end{figure}

Figure~\ref{fig:real-decoders} shows the results. \textbf{When calibrated, the product decoder was more accurate}: by $+0.3$ to $+0.4$ points for all codes (95\% bootstrap confidence intervals excluding $0$), and significantly better by McNemar's test in 24 of the 39 cells (dataset, $M$, code), the sum decoder in none. \textbf{Uniform overconfidence widens the product advantage} ($+0.9$ to $+1.8$ points at $\gamma=4$). The product decision, a signed sum of logits (Eq.~\eqref{eq:pair-diff}), is unchanged by a common factor. The sum decoder, in contrast, moves toward a Hamming-distance majority vote as the sigmoid saturates, which is consistent with its loss of accuracy. \textbf{Calibration error varying across columns reverses the sign.} At $\sigma=1$ the sum decoder was better by $0.9$ to $2.0$ points, significantly in 36 of the 39 cells. This is consistent with how the decoders weight the votes: the product decoder weights confident columns heavily (Section~\ref{sec:votes}), so overconfidence in a few columns can dominate its decision, whereas the sum votes are bounded in $[-1,1]$. We did not model how the column errors combine; Section~\ref{sec:discussion} lists this as future work. The reversal was largest on Hadamard codes.

The estimated margin at disagreement points (the gap between the top two sum scores) was small, about 7\% of the overall mean (median), and below the overall mean in all 139 of the 156 (dataset, $M$, code, $\gamma$) conditions with at least 20 disagreements: disagreements concentrate near ties.

\section{Discussion}\label{sec:discussion}

\subsection{Which decoder should be used?}

By the criterion of noiseless consistency the answer is clear. Exact MAP-consistency requires an equidistant code for the sum decoder (Proposition~\ref{prop:sum-equidistant}) and a code with pair structure, that is, a union of $n$vR codes, for the product decoder (Theorem~\ref{th:product-pair}), and pair structure implies equidistance (Corollary~\ref{cor:product-implies-sum}). The sum decoder is thus safe on a wider class of codes. In particular, the Hadamard code is an efficient equidistant code with distance $M/2$ and length $M-1$, but for $M\geq8$ it has no pair structure (Proposition~\ref{prop:hadamard-no-pair}); with a Hadamard code, choose the sum decoder if exact consistency is needed. The $n$vR codes are consistent for both decoders, but column deletion breaks pair structure and equidistance (Experiment~A), so the code must be used in full.

Product reversals matter only in a limited range. $\bar\delta\approx0.018$ is a proven \emph{lower} bound showing that the reversal margin can be this large (Theorem~\ref{th:worst-case}), not an upper bound guaranteeing safety; local search has found reversals up to $0.025$. Under random posteriors the MAP margin at disagreement points was at most about half of the construction value (Experiment~B); either way, reversals are confined to near ties, where the posteriors are nearly equal.

Under estimation error the picture changes (Experiment~C). With calibrated logistic regression the product decoder is slightly more accurate, and its decision is unchanged under uniform overconfidence of all columns; when the calibration error differs across columns, the sum decoder is better. This is consistent with the product decoder reading the votes as unbounded log-odds and the sum decoder as bounded probabilities. In practice, use the product decoder if the probability of each column can be calibrated individually, and the sum decoder if the calibration quality varies across columns (or cannot be checked). Either way the difference is within about 2 points (0.3--0.4 points when calibrated), and disagreements are confined to near ties. The claim of \citet{kittler1998combining} that the sum is robust to estimation error can be read as robustness to calibration error varying across columns.

\subsection{Limitations and future work}

Our theory assumes the noiseless setting $\bm p=C^\top\vpi$, which separates the decoding rule and the code structure from classifier error but does not hold for real classifiers: under estimated probabilities, even 2vR, on which both decoders agree with MAP in the noiseless setting, yields different sum and product decisions (Experiment~C). How robust equidistance and pair structure are to estimation error is unknown. Experiment~C suggests that, under estimation error, which decoder wins may depend on how the column errors combine; analyzing this through error exponents (reliability functions) is future work.

For the reversal margin of Hadamard codes, open problems are (1)~the exact worst MAP margin with argmax disagreement (at least $0.0166$; local search found at most $0.025$) and an analytic derivation of the threshold $0.911\,\bar\delta$ in the construction (currently only numerical); (2)~the lower-bound constant for non-Sylvester types (numerically, a similar $\Theta(10^{-2})$ is observed); and (3)~the exact rate of decay in the typical case.

\section{Conclusion}\label{sec:conclusion}

We have characterized the MAP-consistency of the product decoder in ECOC under the noiseless setting $\bm p=C^\top\vpi$. Reading a class-pair comparison as a sample of votes, the sum decoder decides by the sample mean and the product decoder by an M-estimator through $\artanh$, and for every code each vote is the MAP margin plus a bias determined by the background alone. The sum decoder is MAP-consistent if and only if the code is equidistant \citep{james1998error} (Proposition~\ref{prop:sum-equidistant}). Our main theorem (Theorem~\ref{th:product-pair}): the product decoder is MAP-consistent if and only if every class pair has pair structure, which strengthens the sufficient symmetry condition of prior work \citep{kumoi2021IPSJ} to a necessary and sufficient one, and if and only if the code is a union of $n$vR codes. Every product-consistent code is sum-consistent (Corollary~\ref{cor:product-implies-sum}). Hadamard codes ($M\geq8$), which meet the Plotkin bound, have no pair structure because their columns are orthogonal (Proposition~\ref{prop:hadamard-no-pair}), and for the Sylvester type the reversal margin can be at least a constant $\bar\delta\approx0.018175$ independent of $M$ (Theorem~\ref{th:worst-case}). For $M$ not a multiple of four, row deletion gives equidistant codes at a cost of at most three extra columns.

Experiments confirmed Proposition~\ref{prop:sum-equidistant} on the codes of Experiment~A and the main theorem on 977 class pairs. Under estimation error, the product decoder was slightly more accurate when each column was calibrated, and the sum decoder was better when the calibration error differed across columns; with real classifiers it appears safe to choose the product decoder if the calibration of each column can be checked, and the sum decoder otherwise. The core difference between the two decoders is that the sum decoder requires only first-moment balance of the votes (equidistance), whereas the product decoder requires symmetry of the whole vote distribution (pair structure), and this symmetry leaves only the $n$vR codes.

\ifreview\else
\subsubsection*{Acknowledgments}
The author is deeply grateful to Shigeichi Hirasawa, Manabu Kobayashi, and Hideki Yagi for many years of seminar discussions; this paper is the culmination of the discussions that began with the earlier work \citep{kumoi2021IPSJ}. This work was supported by JSPS KAKENHI Grant Number JP24K20833.
\fi

\bibliography{ref}

\begin{thebibliography}{16}
\providecommand{\natexlab}[1]{#1}
\providecommand{\url}[1]{\texttt{#1}}
\expandafter\ifx\csname urlstyle\endcsname\relax
  \providecommand{\doi}[1]{doi: #1}\else
  \providecommand{\doi}{doi: \begingroup \urlstyle{rm}\Url}\fi

\bibitem[Allwein et~al.(2000)Allwein, Schapire, and
  Singer]{allwein2000reducing}
Erin~L. Allwein, Robert~E. Schapire, and Yoram Singer.
\newblock Reducing multiclass to binary: A unifying approach for margin
  classifiers.
\newblock \emph{Journal of Machine Learning Research}, 1:\penalty0 113--141,
  2000.

\bibitem[{\'A}lvarez-P{\'e}rez \& Callejas-Ramos(2026){\'A}lvarez-P{\'e}rez and
  Callejas-Ramos]{alvarez2026walsh}
Lorena {\'A}lvarez-P{\'e}rez and {\'A}lvaro Callejas-Ramos.
\newblock Walsh sequences as direct error correcting output code dichotomies
  for multiclass problems.
\newblock \emph{Engineering Applications of Artificial Intelligence},
  173:\penalty0 114442, 2026.
\newblock \doi{10.1016/j.engappai.2026.114442}.

\bibitem[DeGroot \& Mortera(1991)DeGroot and Mortera]{degroot1991optimal}
Morris~H. DeGroot and Julia Mortera.
\newblock Optimal linear opinion pools.
\newblock \emph{Management Science}, 37\penalty0 (5):\penalty0 546--558, 1991.
\newblock \doi{10.1287/mnsc.37.5.546}.

\bibitem[Dietterich \& Bakiri(1995)Dietterich and
  Bakiri]{dietterich1995solving}
Thomas~G. Dietterich and Ghulum Bakiri.
\newblock Solving multiclass learning problems via error-correcting output
  codes.
\newblock \emph{Journal of Artificial Intelligence Research}, 2:\penalty0
  263--286, 1995.
\newblock \doi{10.1613/jair.105}.

\bibitem[Genest \& Zidek(1986)Genest and Zidek]{genest1986combining}
Christian Genest and James~V. Zidek.
\newblock Combining probability distributions: A critique and an annotated
  bibliography.
\newblock \emph{Statistical Science}, 1\penalty0 (1):\penalty0 114--135, 1986.
\newblock \doi{10.1214/ss/1177013825}.

\bibitem[James \& Hastie(1998{\natexlab{a}})James and Hastie]{james1997error}
Gareth James and Trevor Hastie.
\newblock The error coding and substitution pacts.
\newblock In \emph{Advances in Neural Information Processing Systems 10 (NIPS
  1997)}, pp.\  542--548, 1998{\natexlab{a}}.

\bibitem[James \& Hastie(1998{\natexlab{b}})James and Hastie]{james1998error}
Gareth James and Trevor Hastie.
\newblock The error coding method and {PICTs}.
\newblock \emph{Journal of Computational and Graphical Statistics}, 7\penalty0
  (3):\penalty0 377--387, 1998{\natexlab{b}}.
\newblock \doi{10.1080/10618600.1998.10474782}.

\bibitem[Kittler et~al.(1998)Kittler, Hatef, Duin, and
  Matas]{kittler1998combining}
Josef Kittler, Mohamad Hatef, Robert P.~W. Duin, and Ji{\v{r}}{\'\i} Matas.
\newblock On combining classifiers.
\newblock \emph{IEEE Transactions on Pattern Analysis and Machine
  Intelligence}, 20\penalty0 (3):\penalty0 226--239, 1998.
\newblock \doi{10.1109/34.667881}.

\bibitem[Kumoi et~al.(2021)Kumoi, Yagi, Kobayashi, Goto, and
  Hirasawa]{kumoi2021IPSJ}
Gendo Kumoi, Hideki Yagi, Manabu Kobayashi, Masayuki Goto, and Shigeichi
  Hirasawa.
\newblock A study on the optimization of the {ECOC} method for multi-label
  classification problems (in {J}apanese).
\newblock \emph{IPSJ Transactions on Mathematical Modeling and Its Applications
  (TOM)}, 14\penalty0 (3):\penalty0 1--10, 2021.
\newblock URL \url{https://ipsj.ixsq.nii.ac.jp/records/212227}.

\bibitem[Kumoi et~al.(2023)Kumoi, Yagi, Kobayashi, Goto, and
  Hirasawa]{kumoi2023IJNS}
Gendo Kumoi, Hideki Yagi, Manabu Kobayashi, Masayuki Goto, and Shigeichi
  Hirasawa.
\newblock Performance evaluation of error-correcting output coding based on
  noisy and noiseless binary classifiers.
\newblock \emph{International Journal of Neural Systems}, 33\penalty0
  (2):\penalty0 2350004, 2023.
\newblock \doi{10.1142/S0129065723500041}.

\bibitem[Passerini et~al.(2004)Passerini, Pontil, and
  Frasconi]{passerini2004new}
Andrea Passerini, Massimiliano Pontil, and Paolo Frasconi.
\newblock New results on error correcting output codes of kernel machines.
\newblock \emph{IEEE Transactions on Neural Networks}, 15\penalty0
  (1):\penalty0 45--54, 2004.
\newblock \doi{10.1109/TNN.2003.820841}.

\bibitem[Plotkin(1960)]{plotkin1960binary}
Morris Plotkin.
\newblock Binary codes with specified minimum distance.
\newblock \emph{IRE Transactions on Information Theory}, 6\penalty0
  (4):\penalty0 445--450, 1960.
\newblock \doi{10.1109/TIT.1960.1057584}.

\bibitem[Ramaswamy et~al.(2014)Ramaswamy, Srinivasan~Babu, Agarwal, and
  Williamson]{ramaswamy2014consistency}
Harish~G. Ramaswamy, Balaji Srinivasan~Babu, Shivani Agarwal, and Robert~C.
  Williamson.
\newblock On the consistency of output code based learning algorithms for
  multiclass learning problems.
\newblock In \emph{Proceedings of the 27th Conference on Learning Theory
  (COLT)}, volume~35 of \emph{PMLR}, pp.\  885--902, 2014.

\bibitem[Tax et~al.(2000)Tax, van Breukelen, Duin, and
  Kittler]{tax2000combining}
David M.~J. Tax, Martijn van Breukelen, Robert P.~W. Duin, and Josef Kittler.
\newblock Combining multiple classifiers by averaging or by multiplying?
\newblock \emph{Pattern Recognition}, 33\penalty0 (9):\penalty0 1475--1485,
  2000.
\newblock \doi{10.1016/S0031-3203(99)00138-7}.

\bibitem[Verma \& Swami(2019)Verma and Swami]{verma2019error}
Gunjan Verma and Ananthram Swami.
\newblock Error correcting output codes improve probability estimation and
  adversarial robustness of deep neural networks.
\newblock In \emph{Advances in Neural Information Processing Systems 32
  (NeurIPS 2019)}, 2019.

\bibitem[Yin et~al.(2005)Yin, Xie, and Kuang]{Yin2005-nu}
An-Rong Yin, Xiang Xie, and Jingming Kuang.
\newblock Using {Hadamard} {ECOC} in multi-class problems based on {SVM}.
\newblock In \emph{Proceedings of Interspeech 2005}, pp.\  3125--3128, 2005.
\newblock \doi{10.21437/Interspeech.2005-672}.

\end{thebibliography}
\bibliographystyle{tmlr}

\appendix
\section{Proofs from the Main Text}\label{app:proofs}

\paragraph{Proof of Proposition~\ref{prop:pair-inversion}.}
When $\vpi$ is in the relative interior, $0<q_j<1$, so $\beta+a_j=z_j\in(-1,1)$ and $\Psi_{\bm b}'(\beta)=\sum_{j\in T}(1-(\beta+a_j)^2)^{-1}>0$. $\Psi_{\bm b}$ is defined on the open interval $B=(-1-\min_ja_j,\,1-\max_ja_j)$ on which $\beta+a_j\in(-1,1)$ for all $j$, and it diverges to $-\infty$ at the left endpoint and $+\infty$ at the right endpoint, so it has a unique root on $B$. Since $|a_j|\leq\sum_kb_k=1-r$, we have $(-r,r)\subset B$.

The product decoder prefers $m$ iff $\Psi_{\bm b}(\beta)>0$, and MAP prefers $m$ iff $\beta>0$. (1)~If $\Psi_{\bm b}(0)=0$, monotonicity gives $\operatorname{sign}\Psi_{\bm b}(\beta)=\operatorname{sign}\beta$. (2)~If $\Psi_{\bm b}(0)>0$, the root lies at $\beta^*<0$; for $\beta\in(\beta^*,0)$ the reversal $\Psi_{\bm b}(\beta)>0>\beta$ occurs, and for $\beta\notin[\beta^*,0]$ the signs agree. Since $\beta$ on the family is restricted to $(-r,r)$, the reversal interval is $(\max(\beta^*,-r),0)$, which is nonempty because $\max(\beta^*,-r)<0$.

\paragraph{Proof of Theorem~\ref{th:worst-case}.}
Fix any class $m\neq0$ and take the class pair $(0,m)$. Since $\bm c_0=\bm1$, $T=\{j:c_{mj}=0\}$ and $|T|=M/2$; in the character form of Proposition~\ref{prop:invisible-class}, $T=\{j:\chi_m(j)=-1\}$, and the agreement sign of a background class $k$ is $\tau_{kj}=2c_{kj}-1=\chi_k(j)$. Choose linearly independent $b_1,b_2\in\mathbb F_2^t\setminus\{0,m\}$ with $b_1\oplus b_2\neq m$ (possible for $M\geq8$) and let $b_3:=b_1\oplus b_2$. Place background mass $\rho=1/5$ on each element of the triple and set the other background classes to $0$. Then $\pi_0+\pi_m=2/5$ and $\beta=\pi_0-\pi_m$.

By Lemma~\ref{lem:vote-decomp}, the bias is $a_j=\rho\,(\chi_{b_1}(j)+\chi_{b_2}(j)+\chi_{b_3}(j))$. By the product rule $\chi_{b_3}=\chi_{b_1}\chi_{b_2}$, $\chi_{b_1}+\chi_{b_2}+\chi_{b_3}=(1+\chi_{b_1})(1+\chi_{b_2})-1\in\{3,-1\}$, taking the value $3$ only when $\chi_{b_1}=\chi_{b_2}=1$. Since $m,b_1,b_2$ are linearly independent, $(\chi_{b_1},\chi_{b_2})$ is uniformly distributed on $\{\pm1\}^2$ over $T=\{\chi_m=-1\}$. Hence
$$a_j\in\{+0.6\ (\text{proportion}\ 1/4),\ -0.2\ (\text{proportion}\ 3/4)\},\qquad\langle a\rangle=0,$$
and
$\Psi_{\bm b}(\beta)=\tfrac s4\artanh(\beta+0.6)+\tfrac{3s}4\artanh(\beta-0.2)$, $\Psi_{\bm b}(0)=s\,[\tfrac14\artanh0.6+\tfrac34\artanh(-0.2)]\approx0.0212\,s>0$. By Proposition~\ref{prop:pair-inversion}(2) there is a root $\beta^*<0$. Writing $\beta^*=-\delta$ and using the odd symmetry of $\artanh$, the root equation becomes $\artanh(0.6-\delta)=3\artanh(0.2+\delta)$, i.e., Eq.~\eqref{eq:worst-case-eq}, which does not involve $s$. Since $r=2/5>\bar\delta$, the reversal interval is $\beta\in(-\bar\delta,0)$, i.e., $0<\pi_m-\pi_0<\bar\delta$. Then $\pi_m=(2/5+\varepsilon)/2>1/5$ exceeds the triple masses $1/5$ and $\pi_0$, so $m$ is the MAP class.

This construction has mass on only five classes and lies on the boundary of the simplex. Distribute a small mass $\eta>0$ uniformly over the remaining $M-5$ classes and scale the five masses by $1-(M-5)\eta$ to obtain $\vpi_\eta$. Each $p_j$ on $T$ is continuous in $\eta$, and at $\eta=0$ the values $p_j\in\{0.4,0.8\}$ are bounded away from the endpoints. For fixed $\varepsilon\in(0,\bar\delta)$ the reversal and the fact that $m$ is the MAP class hold at $\eta=0$ with strict inequalities, so they persist for all sufficiently small $\eta>0$ (the scaling turns $\pi_m-\pi_0$ into $\varepsilon(1-(M-5)\eta)$; starting from $\varepsilon/(1-(M-5)\eta)$ restores $\varepsilon$).

\section{\texorpdfstring{Proof of (2)$\Rightarrow$(3) in the Main Theorem}{Proof of (2) => (3) in the Main Theorem}}\label{app:proof-product}

Suppose some class pair $(m,m')$ lacks pair structure, and assume weak consistency; we derive a contradiction. Let $U:=\{\bm b:b_k>0,\ \sum_kb_k<1\}$ be the set of admissible backgrounds.

\textbf{Step 1.} $\Psi_{\bm b}(0)=0$ for every $\bm b\in U$. Indeed, if $\Psi_{\bm b}(0)>0$ for some $\bm b\in U$, Proposition~\ref{prop:pair-inversion}(2) (and, if $\Psi_{\bm b}(0)<0$, the same proposition with $m\leftrightarrow m'$) gives a strict reversal of the MAP order on a nonempty interval of $\beta$, contradicting weak consistency.

\textbf{Step 2 (a single ray).} Number the background classes $k_1,\ldots,k_{M-2}$ and set $v_{k_i}:=2^{i-1}/2^M$. Since $v_k>0$ and $\sum_kv_k<1$, we have $t\bm v\in U$ for $t\in(0,1)$, and Step 1 gives
$$\sum_{j\in T}\artanh(t\,u_j)=0,\qquad u_j:=\langle\bm v,\bm\tau_j\rangle\in(-1,1).$$
For $|t|<1$ the left side equals the convergent power series $\sum_{\kappa\,\text{odd}}\frac{t^\kappa}{\kappa}\sum_{j\in T}u_j^\kappa$, an odd function of $t$. It therefore vanishes on all of $(-1,1)$, and uniqueness of power-series coefficients gives $\sum_{j\in T}u_j^\kappa=0$ for every odd $\kappa$.

\textbf{Step 3 (symmetry).} Let $S$ be the finite set of the values $\pm u_j$ and $\nu(x):=\#\{j\in T:u_j=x\}-\#\{j\in T:u_j=-x\}$. For odd $\kappa$, $\sum_{x\in S}\nu(x)\,x^\kappa=2\sum_{j\in T}u_j^\kappa=0$; for even $\kappa$ the two terms cancel. These equations for $\kappa=0,1,\ldots,|S|-1$ imply $\nu\equiv0$ by the nonsingularity of the Vandermonde matrix, i.e., the multiset $\{u_j\}_{j\in T}$ is symmetric about $0$.

\textbf{Step 4 (back to patterns).} With $A_j$ the set of background classes with $\tau_{kj}=+1$, $u_j=2^{-M}\bigl(2\sum_{k_i\in A_j}2^{i-1}-(2^{M-2}-1)\bigr)$. By uniqueness of binary expansions, $u_j$ determines $A_j$ and hence $\bm\tau_j$, and $-u_j$ corresponds to $-\bm\tau_j$. Hence the multiset of agreement patterns is invariant under sign flip, and by Lemma~\ref{lem:pair-pattern} the class pair has pair structure, a contradiction. $\square$

\section{Experimental Details}\label{app:protocol}

Table~\ref{tab:protocol} summarizes the main settings of the experiments in Section~\ref{sec:experiments}.

\begin{table}[tb]
  \centering
  \caption{Experimental settings.}
  \label{tab:protocol}
  \small
  \begin{tabular}{lp{0.85\linewidth}}
    \toprule
    Exp. & Main settings \\
    \midrule
    A & Noiseless agreement: 1-vs-Rest, 2vR, half-vs-half, Hadamard, logM $\times$ $M\in\{4,8,16,32\}$; four series (Dirichlet $\alpha\in\{0.3,1\}$ and two derived from Gaussian data), $200{,}000$ samples each ($100{,}000$ for codes with many columns); column deletion $k=1,2$ from 2vR (10 seeds); pair-structure check on 49 codes and 977 class pairs \\
    A' & Row deletion (Corollary~\ref{cor:subcode}): two rows deleted from Hadamard $M=12$ to give $M=10$ (equal-weight and unequal-weight cases); Dirichlet $\alpha\in\{0.3,1\}$, $50{,}000$ samples each; zero sum-decoder disagreements \\
    B & Hadamard $M\in\{8,12,16,20,28,32,48,64\}$; typical values (largest margin at disagreement points) from Dirichlet $\alpha\in\{0.3,1\}$ ($50{,}000\times3$ seeds) and Gaussian data ($20{,}000\times4$ variances $\times3$ seeds); worst values by multi-start SLSQP (maximum over 100 starts $\times$ 3 seeds); margin distribution at disagreement points from $500{,}000$ noiseless samples for $M\in\{8,16,32\}$ \\
    C & letter, isolet, pendigits, MNIST and Fashion-MNIST (PCA-50, at most 1500 per class), CIFAR-100 (MobileNetV2 features, PCA-100); $M\in\{8,16,24,32\}$ (as far as the number of classes allows); class subsets and splits (train/validation/test $=60/20/20$) varied over 5 seeds; per-column logistic regression (inverse regularization strength $1$, standardized features) with temperature calibration on validation data; (a)~common factor $\gamma\in\{0.5,1,2,4\}$, (b)~column-specific factors $\gamma_j=\exp(\sigma g_j)$, $\sigma\in\{0,0.5,1\}$ (3 draws each). The hypotheses and the design of (a) were fixed before the runs \\
    \bottomrule
  \end{tabular}
\end{table}

\paragraph{Reproducibility.}
\ifreview The Lean~4 formalization of all theorems, lemmas, propositions, and corollaries is provided as supplementary material and will be released in a public repository after acceptance.\else The Lean~4 formalization of all theorems, lemmas, propositions, and corollaries is provided in the ancillary files of this arXiv submission.\fi{} The numerical claims are confirmed by verification scripts, and the code to reproduce the experiments will be released in a public repository after acceptance. The product score is computed as the sum of the Bernoulli log-likelihoods of the binary-classifier outputs (Eq.~\eqref{eq:product-score}), i.e., the exact log-likelihood rule rather than a $\log(1+p)$-type approximation. CIFAR-100 features are extracted once and cached for reuse.

\end{document}